\documentclass{article}
\usepackage{spconf,amsmath,graphicx}
\usepackage{epsfig}
\usepackage{amssymb}
\usepackage{subfig}
\usepackage{cite}
\usepackage{multirow}
\usepackage{epstopdf}
\usepackage{graphicx}
\usepackage{amsmath}
\usepackage{color, soul}
\usepackage[style=base]{caption}
\usepackage{lineno}
\usepackage{comment}
\usepackage{array}  
\usepackage{siunitx}
\usepackage{mathtools}
\usepackage{calrsfs}
\usepackage{eucal}
\usepackage{bm}
\newcolumntype{d}{>{\displaystyle}c}

\usepackage{tabularx}

\usepackage{subfig}

\usepackage{titlesec}

\titleformat{\paragraph}
{\normalfont\normalsize\bfseries}{\theparagraph}{1em}{}
\titlespacing*{\paragraph}
{0pt}{3.25ex plus 1ex minus .2ex}{1.5ex plus .2ex}

\usepackage{url}
\usepackage{hyperref}
\hypersetup{
	colorlinks=true,       
	linkcolor=red,          
	citecolor=green,        
	filecolor=magenta,      
	urlcolor=cyan           
}
\usepackage{enumitem}
\usepackage{pgfplots}

\usepackage{array}
\newcolumntype{L}[1]{>{\raggedright\let\newline\\\arraybackslash\hspace{0pt}}m{#1}}
\newcolumntype{C}[1]{>{\centering\let\newline\\\arraybackslash\hspace{0pt}}m{#1}}
\newcolumntype{R}[1]{>{\raggedleft\let\newline\\\arraybackslash\hspace{0pt}}m{#1}}

\title{A survey of top-down approaches for human pose estimation}
\name{
	\begin{tabular}{ccc}
		 Thong Duy Nguyen & Milan Kresovi\'c 
	\end{tabular}
}

\address{
	\begin{tabular}{c}
		Norwegian University of Science and Technology, Norway.
	\end{tabular}
}

\begin{document}
%
\maketitle

\begin{abstract}
Human pose estimation in two-dimensional images videos has been a hot topic in the computer vision problem recently due to its vast benefits and potential applications for improving human life, such as behaviors recognition, motion capture and augmented reality, training robots, and movement tracking. Many state-of-the-art methods implemented with Deep Learning have addressed several challenges and brought tremendous remarkable results in the field of human pose estimation. Approaches are classified into two kinds: the two-step framework (top-down approach) and the part-based framework (bottom-up approach). While the two-step framework first incorporates a person detector and then estimates the pose within each box independently, detecting all body parts in the image and associating parts belonging to distinct persons is conducted in the part-based framework. This paper aims to provide newcomers with an extensive review of deep learning methods-based 2D images for recognizing the pose of people, which only focuses on top-down approaches since 2016. The discussion through this paper presents significant detectors and estimators depending on mathematical background, the challenges and limitations, benchmark datasets, evaluation metrics, and comparison between methods.
\end{abstract}

%


\section{Introduction:}

Vision-based human pose estimation refers to the classification and localization of human keypoints in a given frame and connecting the corresponding keypoints for approximating the skeleton. The keypoints are subjective and vary from application to application. However, a joint that characterizes the human body shape can be defined a keypoint. For example, hinge joints, pivot joints, ellipsoidal joints and the ball and socket joint that connect the shoulder and hip joints to allow backward, forward, sideways, and rotating movements are considered necessary keypoints included in most of the pose estimation algorithms. It is one of the classical problems in computer vision and stayed an active field of research for decades. As the field evolved, the application spectrum also widened, and new innovative products incorporated it into their products and service. For example, human pose estimation plays a vital role in virtual and augmented reality \cite{su2019deep, ullah2021social, schmidtke2021unsupervised}. It is an essential ingredient in many multimedia applications related to entertainment, like gaming \cite{assadzadeh2022vision, pardos2022unifying} and animated movies. It plays a critical role in athletes' performance analysis \cite{ullah2021attention, araujo2021artificial} and training in sports \cite{khan2019person, wang2019ai}. However, the applications are not limited to this and play an important role in other domains like medical imaging \cite{mcnally2020evopose2d, bilal2019chemometric, casas2019patient}, crowd analysis \cite{ullah2021multi,  choi20213dcrowdnet, ullah2017density ,riaz2021anomalous, ullah2016crowd, stenum2021two, ullah2019two, khan2020crowd, ullah2018anomalous, khan2019disam}, animal farming \cite{fang2021pose, quddus2020bottom}, action recognition \cite{song2021human, ullah2017human}, human-computer interaction \cite{moon2018v2v, chua2022hand}, detection \cite{hofer2021object,khan2020tcm, PedAppearance, chen2021monorun, varamesh2020mixture, hagelskjaer2020pointvotenet} and tracking \cite{shagdar2021geometric, ullah2020head, chen2020cross,  ullah2019single, wang2020combining, 58, zhou2020temporal, 59, ariz2016novel,  ullah2018deep, chu2021part, 61, zhang2021voxeltrack,ullah2019siamese} specifically and video surveillance \cite{10, 55} in general, segmentation \cite{zhuang2021semantic,ullah2019hybrid, zhai2021jd, ullah2018pednet}, autonomous driving \cite{zhao2021shape, ding2021globally, gu2019efficient}, behaviour analysis \cite{fang2021pose, ullah2019stacked}, facial emotion recognition \cite{bisogni2021ifepe, alreshidi2020facial}, and the gait analysis \cite{lonini2022video, stenum2021two}. 

In short, human pose estimation is an important topic and has enormous potential in the indusdry of varied nature.   
Several models have been proposed to handle this task early, such as the Pictorial Structures (PS) model of Fischler and Elschlager \cite{fischler1973representation}, Deformable Part Model (DPM) of Felzenswalb \cite{felzenszwalb2008discriminatively}. These models have a tree-structured graphical type and perform good results conditioned in visible limbs. However, they fail if incorrectly capturing the correlation between variables happens with the tree-structured. With the development of deep learning in the scientific community nowadays. Many state-of-the-art methods yielded excellent results and solved several challenges classical methods could not handle. For example, the RMPE method \cite{fang2017rmpe} proposed a framework for estimation in inaccurate human bounding boxes. The framework is consisted of three primary components: a Symmetric Spatial Transformer Network (SSTN), Parametric Pose NonMaximum-Suppression (NMS), and a Pose-Guided Proposals Generator (PGPG). Another method is CPN \cite{chen2018cascaded}, which aims to relieve the problem from “hard” key points (occluded keypoints, invisible keypoints and complex background) by applying two stages algorithms: GlobalNet and refinement.

\begin{figure*}
	\centering
	\includegraphics[scale=0.8]{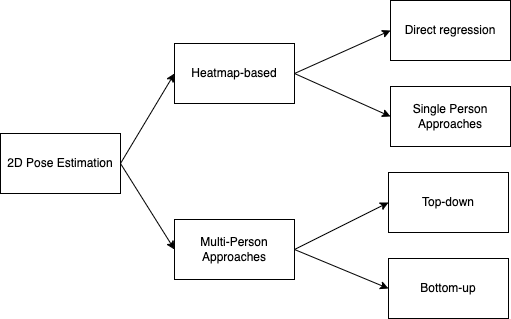}
	\caption{Taxonomy of Deep Learning-based method for 2D pose estimation. \cite{dang2019deep}} 
	\label{fig1}
\end{figure*}

The umbrella human pose estimation is categorized into two types of methodologies \cite{dang2019deep}, as shown in figure \ref{fig1}: single-person pose and multi-person pose estimation approaches. On the one hand, single-person methods aim to solve the regression problem by locating the human parts of the person assumed to dominate the image content, such as the left/right ears, the center of the neck, and the left/right shoulders. On the other hand, the multi-person approaches address the unconstrained problem since the image's unknown number and are position are presented. 

\begin{figure}[!htp]
	\centering
	\includegraphics[scale=0.3]{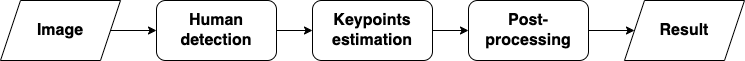}
	\caption{The framework of the top-down pipeline. } 
	\label{fig2}
\end{figure}

In single-person pose estimation, the pipeline is divided into two types based on predicting vital points \cite{dang2019deep}: direct regression-based approaches and heatmap-based approaches. While the first one utilizes the output feature maps to regress key points directly, the second predicts points from earlier generated heatmaps. Multi-person methodologies are classified into bottom-up and top-down approaches. In bottom-up methods, the first step is to predict all the critical points before associating them with the person they belong to. As shown in figure \ref{fig2}, the workflow of top-down approaches has similar steps with a reversed order, which starts detecting and locating persons in single bounding boxes, then estimates the pose.

In summary, this paper only reviews deep learning-based methods on 2D images conducted in the field of top-down approaches. Based on two top-down approaches, the main content is presented in two parts: the first part provides a summary of recent significant human detection models and the second part analyzes human pose estimators. Notably, the information written in these parts is about mathematical backgrounds, evaluation metrics, and main workflows. The rest of the paper discusses benchmark datasets, methods comparison, and limitations of each model.

This review paper is organized as follows. Section \ref{sec:related} shortly discusses the related work in the field. The key component of top down approaches i.e. object detection is explained in Section \ref{sec:obj}. The state-of-the-art top-down approaches is discussed in section \ref{sec:TDP}. The prominent evaluation metrics are listed in section in section \ref{sec:EXP} and final remarks are given in section \ref{sec:CON} that concludes the papers. 

\section{Related work} 
\label{sec:related}
Many surveys related to human pose estimation problems have been published annually. However, most of them have concentrated on conventional methodologies and rarely provided information about deep learning-based methods. Several latest papers extensively surveyed models built from deep learning for predicting human key points but in large scopes. For instance, \cite{munea2020progress} discussed different deep learning-based 2D human pose estimation models but covered both bottom-up and top-down approaches. \cite{chen2020monocular} presented the review in the current deep learning methods in 2D and 3D and two-step and part-based frameworks. This survey just shortly summarized related concepts and mainly focused on dataset and evaluation protocols comparison. Also, the mathematical parts were not analyzed deeply. Another survey on 2D human pose estimation focusing on the deep learning field is \cite{dang2019deep}. This survey started by presenting the taxonomy of pose estimation pipelines in single-pose and multi-pose approaches, followed by presenting detail of each estimation algorithm, such as direct regression and heatmap. However, state-of-the-art human detection models were not mentioned through this survey in term of top-down methods.

Our paper is scaled down the research topic focusing on only one approach. In other words, this paper surveys recent top-down methods. However, the discussion through every process is in-depth analyzed for each detection and estimation part.

\section{Object Detection}
\label{sec:obj} 
Object detection is a key computer vision problem, which is the progress of identifying multiple instances of visual objects in digital images, then localizing the position of objects. Over many years, object detection has been expanded and explored to different technical breakthroughs: edge detection, human face detection, pose prediction, and pedestrian's behaviors detection, etc. Moreover, object detection has been widely applied to human life nowadays such as human tracking, autonomous driving, medical treatments, security fields etc. Regarding top-down human pose estimation approaches, object detection is the first step of the top-down framework to detect multi-person in a scene and localize each person within a rectangle called a bounding box. Generally, modern object detectors are separated into CNN based two-stage detectors and CNN based one-stage detectors. The former one is renowned with the high localization and object recognition accuracy such as Fast R-CNN \cite{girshick2014rich}, Faster R-CNN \cite{ren2015faster}. The latter one is more powerful in real-time running speed, for example, SSD \cite{liu2016ssd}. Take Faster R-CNN as an example for two-stage detectors, the first stage is to propose candidate object bounding boxes by using RPN (Region Proposal Network) and the second one operates RoIPool (Region of Interest Pooling layer) to extract features from each output box in order to perform the following classification and bounding-box regression tasks. By contrast, due to skipping region proposal step and directly proposing predicted boxes from input images, the one-stage detectors have simpler architectures being efficient for real-time devices. 
\subsection{Backbone networks} 
The backbone is the convolutional neural networks (CNN), which is used for filtering significant features by taking images as input and producing feature maps of the corresponding input image. These networks are mainly used for classification task. Object classification problems aims to categorize a single object in the image, and they output a one-hot vector consisting of probabilities of each training class based on the last fully connected layers. In contrast, the object detection problem is more complicated because it requires the model be able to predict several objects and localize the positions of them in a single image. 
Depending on the goal of the project, the backbone can be categorized as deeper and densely, or lightweight networks. The deeper and densely ones are for those who target to achieve the accuracy such as ResNet, ResNeXt, VGG, meanwhile the lightweight ones tend to focus on the efficiency like MobileNet, MobileNetV2. Furthermore, for high precision and accuracy, the complexity of deeper and densely backbone networks are needed. On the other hand, to obtain the smooth and flexibility in running time as well as significant accuracy for video or webcam, the backbone architecture needs to be well-design in order to make a trade-off between speed and accuracy. Backbone networks in object detection play a crucial role. The more correct features are extracted, the higher accuracy objects are detected. \newline
\subsection{CNN based two-stage detectors}
\begin{figure}[!htp]
	\centering
	\includegraphics[scale=0.45]{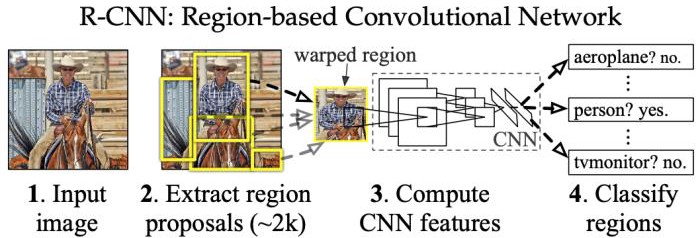}
	\caption{R-CNN overview: The system (1) takes an input image, (2) extracts region proposals, (3) computes features for each proposal, and then (4) classifies each region. \cite{girshick2014rich}.} 
	\label{fig3}
\end{figure}

\textbf{R-CNN (Region-based Convolutional Network):} \newline R-CNN is proposed by Girshick \cite{girshick2014rich}, presented as an object detector, using multi-layer convolutional networks to compute highly discriminative, yet invariant, features. With these extracted features, R-CNN classify image regions resulting in detected bounding boxes or pixel-level segmentation masks.
R-CNN object detection system is designed with four components, as shown in Figure \ref{fig3}. The first component produces independent region proposals of each category, being beneficial for defining the set of detection available to the detection system. The second component is built with convolutional network in order to extracts a fixed-length feature vector from each region. The third component is comprised of of class-specific linear SVMs for object classification in a single image. The last one is bounding-box regressor for object localization. 

The model begins with the region search by using selective search method, which is an alternative to exhaustive search in an image to generate region proposals. The selective search basically initializes small regions in an image and merges them with a hierarchical grouping. After receiving region proposals, CNN is applied to extract a fixed-length feature vector with the size being 4096-dimensional. The model requires the region proposal must first convert to the compatible size with the input size of CNN. From the author, the fixed size input is 227 x 227 pixel. Because of the distinct size and aspect ratio of objects presented on images leading to the different size of region proposals in the first step, all pixels in a tight bounding box around it should be warped to the required size. The feature-extracted vector is fed into multiple classifiers to produce probabilities to belong to each class. After that the trained SVM classifier in each class infer a probability to detect the object for a given vector of features. A linear regression is applied to this vector to adapt the shape of bounding box for a region proposal resulting in reducing the localization errors.  
For training, Girshick first pre-train the CNN on a large auxiliary dataset (ImageNet Classification dataset). The last fully connected layer is replaced by the CNN’s ImageNet-specific 100-way classification layer. To adapt CNN with the detection task, SGD (stochastic gradient descent) is used to fine-tune the CNN parameters only on the warped region proposals.  
To define the positive examples and negative examples, the author separated into two cases. The first one is to define the IoU (intersection over union) overlap threshold as 0.5 for the fine-tuning process. Region proposals are negative when the IoU value is below the threshold and vice versa. The second case is when training the SVM classifiers. The positives is between 0.5 and 1 but they are not ground truth, and region proposals have less then 0.3 IoU overlap are negative. When testing with benchmark datasets. The R-CNN model achieved a mAP score of 62.4 percent on the PASCAL VOC 2012 test dataset and a mAP score of 31.4 percent on the 2013 ImageNet dataset. Although RCNN has made a great achievement, the disadvantages from it still obvious: the time-consuming with redundant feature computations made of overlapped proposals leads to an extremely slow detection speed (14s per image with GPU).\newline \newline
\textbf{Fast Region-based Convolutional Network (Fast-RCNN):} \newline Fast R-CNN is the speed-upgraded and higher accuracy version of R-CNN, which is proposed by Girshick in a year later \cite{girshick2015fast}. Because of not sharing computation when passing each region proposal from selective search method to a ConvNet, R-CNN takes a long time on SVMs classification. In contrast, Fast R-CNN takes the entire image as input for feature extraction at the beginning, followed by passing the Region of Interest (RoI) pooling layer to get the fixed required size playing the input of the following classifier and bounding box regressor. All one-time extracted features from the entire image are sent to CNN for classification and localization at the same time. The goal of the Fast R-CNN is to increase the computational time and save more storage disk related to the high number of models necessary to analyze all region proposals in comparison with R-CNN which inputs each region proposals to CNN. On the one hand training R-CNN takes a large amount of time because of the multi-stage process covering pre-training stage, fine-tuning stage, SVMs classification stage and bounding box regression stage. On the other hand, the training progress of Fast R-CNN is a one-stage end-to-end using multi-task loss on each labeled RoI to jointly train the network.
\begin{figure}[!htp]
	\centering
	\includegraphics[scale=0.45]{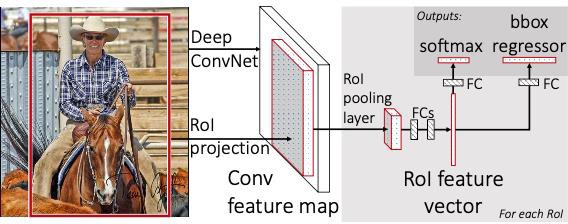}
	\caption{Fast R-CNN architecture: A fully convolutional network is fed an input image and RoIs. FCs pool each RoI into a feature map, which is then mapped to a feature vector. Per RoI, the network produces two output vectors: probabilities distribution by softmax and per-class bounding-box regression offsets. \cite{girshick2015fast}} 
	\label{fig4}
\end{figure}

The framework of Fast R-CNN, as shown in figure \ref{fig4}, basically commences processing the entire image with several conv and max pooling layers to generate a conv feature map. Then, RoI is applied to extract a fixed-length feature vector on each object proposal, from the feature map. The advantage of using RoI for feature extraction is no need of warping regions and reserves the spatial information of features of region proposals. These vectors are passed into a sequence of fully connected layers that is divided into two branches: softmax and bounding box regressor.
Fast R-CNN has made the best mAP scores of 70\%, 68.8\%, and 68.4\% for the test dataset of benchmark 2007 PASLCAL VOC, 2010 PASCAL VOC, and 2012 PASCAL VOC respectively. \newline \newline
\textbf{Faster Region-based Convolutional Network (Faster RCNN):} \newline The Fast R-CNN model has achieved great progress when applying selective search to propose RoI  to reduce the time-consuming but still slow. Faster R-CNN is proposed 3 months later \cite{ren2016faster} to address this issue by replacing the selective method with a novel RPN (Region Proposal Network). This network is a full convolutional network that predicts object bounds and objectness scores at each position very well. The RPN is trained from start to finish to generate high-level object proposals for detection in the Fast R-CNN model. Indeed, Faster R-CNN is a hybrid of RPN and Fast R-CNN.
\begin{figure}[!htp]
	\centering
	\includegraphics[scale=0.45]{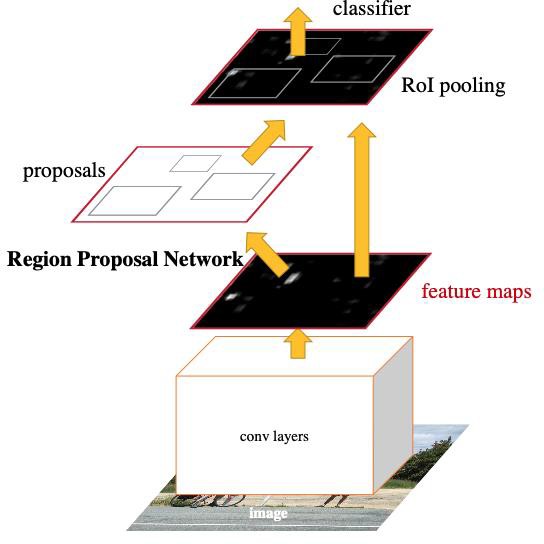}
	\caption{Faster R-CNN architecture. The RPN proposed regions used by the Fast R-CNN later. \cite{ren2016faster}} 
	\label{fig5}
\end{figure}

\begin{figure}[!htp]
	\centering
	\includegraphics[scale=0.45]{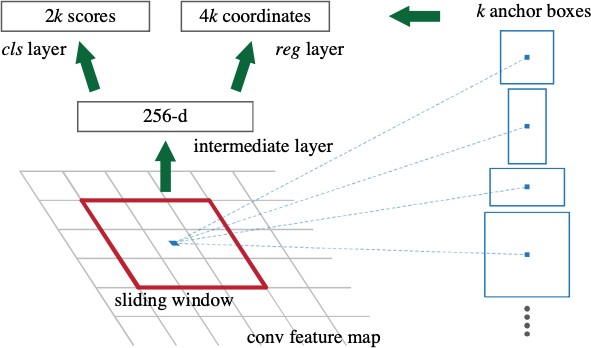}
	\caption{Region Proposal Network (RPN), and anchor boxes in various scales and aspect ratios. \cite{ren2016faster}} 
	\label{fig6}
\end{figure}

The entire framework is illustrated in figure \ref{fig5}. For detailed, the work of generating region proposals is conducted by sliding a small network over the last shared convolutional layer. This small network takes the input as a fixed size window (3 $\times$ 3) . Several boxes with 3 different scales and 3 aspect ratios (figure \ref{fig6}), which are called anchor, are centered at each sliding-window to simplify the proposal generation process with no need of multiple scales of input images or features. On the sliding-window, each center point feature  is relative to a point of the original input image, which is the center point of k ($3\times 3$) anchor boxes.  

\begin{figure}[!htp]
	\centering
	\includegraphics[scale=0.45]{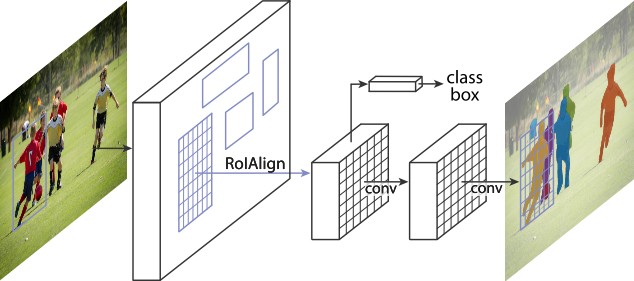}
	\caption{The Mask R-CNN framework for instance segmentation. \cite{he2016deep}} 
	\label{fig7}
\end{figure}

The Faster R-CNN model has been experimented with benchmark dataset such as PASCAL VOC 2007 test set (mAP of 69.9\%). The model's running time (198ms) was nearly 10 times lower than Fast-RCNN's (1830ms) with the same  VGG \cite{simonyan2014very} backbone. \newline \newline
\textbf{Mask Region-based Convolutional Network (Mask R-CNN):} \newline Mask R-CNN \cite{he2017mask} is an object detection model, which is extended from Faster R-CNN by adding one new branch for masks segmentation prediction on RoI. This branch is parallel with other existing branches taking responsibilities for the classification task and bounding box regression. The architecture of Mask R-CNN is composed of two stages (figure \ref{fig7}). The first stage is to propose regions by RPN, and the second stage is based on the proposed feature maps to RoI pool the feature to the region then output the class, bounding box, and binary mask.

The feature extractor is built with ResNet \cite{he2016deep} - FPN (Feature Pyramid Network) backbone to achieve excellent accuracy and processing speed. FPN’s architecture is a bottom-up and top-down structure with lateral connections. In particular, the bottom-up is the CovNet backbone computing feature hierarchy, and the top-down part is to create high-quality features by up-sampling spatially coarser, feature maps from higher pyramid levels. 

\begin{figure}[!htp]
	\centering
	\includegraphics[scale=0.45]{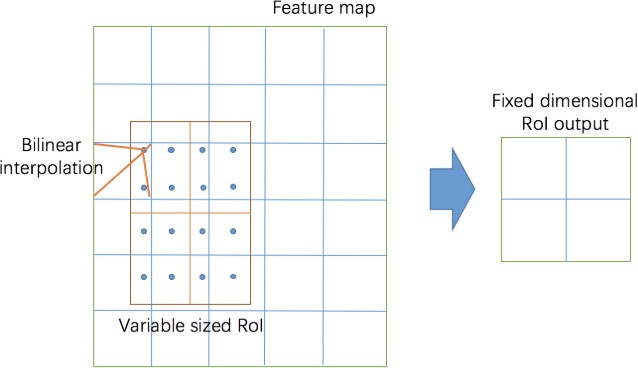}
	\caption{RoIAlign operation. It first computes the floating number coordinates of a feature in the feature map, then uses bilinear interpolation to calculate the exact values of the features in the separated bin. \cite{he2017mask}} 
	\label{fig8}
\end{figure}

Another improvement in the architecture of Mask-RCNN is that the RoIPool is replaced by RoIAlign to extract a small feature map from each RoI. The usual procedure RoIPool quantizes floating-number to the feature map's discrete granularity, then divides it into spatial bins and aggregates the feature values covered by each bin later. As a result, there is a misalignment between the RoI and the extracted features as a result of these quantizations. The RoIAlign replacement addresses this issue by aligning the extracted features with the input. To begin, the author avoids quantizing the RoI boundaries or bins and computes the extracted features using linear interpolation at four regularly sampled locations in each RoI. Finally, to get values of each bin, it aggregates the result (max or average pooling). Read figure \ref{fig8} for more information.  

The Mask R-CNN outperforms all existing, single-model entries on every task, consisting of benchmark dataset COCO 2016 for instance segmentation, bounding box object detection, and person keypoint detection. 
\subsection{CNN based one-stage detectors}
\begin{figure}[!htp]
	\centering
	\includegraphics[scale=0.45]{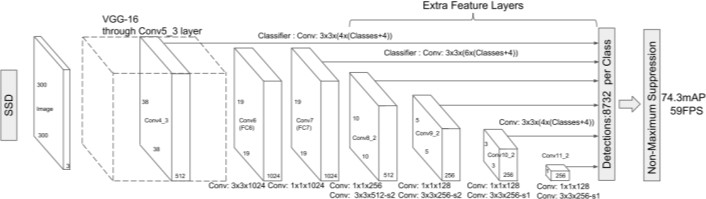}
	\caption{The SSD model extends a base network with several feature layers that predict offsets to default boxes of various scales and aspect ratios, as well as their associated confidences. \cite{liu2016ssd}} 
	\label{fig9}
\end{figure}

\textbf{Single Shot Detector (SSD):} \newline SSD \cite{liu2016ssd} is the second one-stage detector in the deep learning area, developed to predict all bounding boxes and the class probabilities with a end-to-end CNN architecture at a time. This approach, with distinct aspect ratios and scales per feature map location, discretizes the output space of bounding boxes into a set of default boxes. 
The framework of SSD is simple due to removing region proposal generation and subsequent pixel or feature resampling stages, and combining all computational works in a single net-work. In fact, SSD only takes an image as input (figure \ref{fig9}), then passes this image into multi-layer ConvNet with various sizes of filter (10×10,5×5,3×3). After that the bounding boxes are predicted at different locations on the feature maps from convolutional layers. These maps are processed by a specific 3×3 filters to generate a set of bounding boxes similar to anchor boxes in Fast R-CNN model. 

In the network architecture, the early layers, which is called the base network, are the standard backbone network used to classify high-quality images. The rest of the network architecture is auxiliary structures for producing detections, with multi-scale feature maps for detection, convolutional predictors for detection, and default boxes and aspect ratios being the most important. The first is convolutional layers at the end of the truncated base network, which are used to gradually reduce the size of layers for multi-scale detection prediction. The second one is using a set of convolutional filters to generate a fixed set of detection predictions, as the demonstration on the top of the SSD structure in figure \ref{fig9}. The basic filter is $3\times3$ x p small kernel in order to output a class score, or a shape offset relative to the default box coordinates. The last one is described as the offsets prediction being relative to the default box shapes in the cell, and the class scores prediction. 

SSD has achieved great performance with some benchmark dataset in both running speed and detection accuracy such as VOC07 mAP=76.8\%, VOC12 mAP=74.9\%, COCO mAP@.5=46.5\%, mAP@[.5,.95]=26.8\%.

\begin{figure*}[!htp]
	\centering
	\includegraphics[scale=0.85]{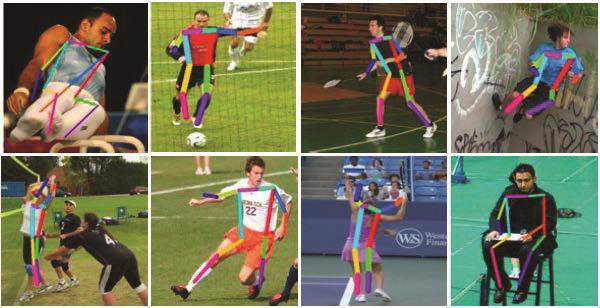}
	\caption{Multi-person estimation poses in top-down approaches.} 
	\label{fig10}
\end{figure*}

\section{2D image Top-down approaches}
\label{sec:TDP}
As mentioned at the beginning in this paper only focuses on comprehensively analyzing several state-of-the-art deep learning-based 2D image in top-down approaches for human pose estimation. The top-down pipeline starts with the detection of individual instances on the given image based on the generated bounding box, and ends with the human pose prediction, as shown in the figure \ref{fig10}. Popular human detection systems are reviewed in the previous session (III. Object detection), which are Fast R-CNN, Faster-RCNN, Mask R-CNN, and SSD. These models are built with various backbone networks as feature extractors, which play an important role for the high accuracy of bounding box regression. The estimation part is surveyed later in this paper, which consists of several significant performance models. In 2016, UC Berkeley proposed a framework called IEF (Iterative Error Feedback) \cite{carreira2016human} to maximize the expensive power of hierarchical feature extractors, such as Convolutional Networks, leading to model rich structure both input and output spaces by introducing top-down feedback. In fact, Convolutional Networks process tasks by using feed-forward architecture, which allows model to learn rich representations of the input space. However, dependencies in the out spaces, which are curial for human pose estimation or object segmentation, is not modelled explicitly. The purpose of proposing the feed-forward architecture is to solve this issue by predicting the error and correct it at the current estimation.
\begin{figure}[!htp]
	\centering
	\includegraphics[scale=0.45]{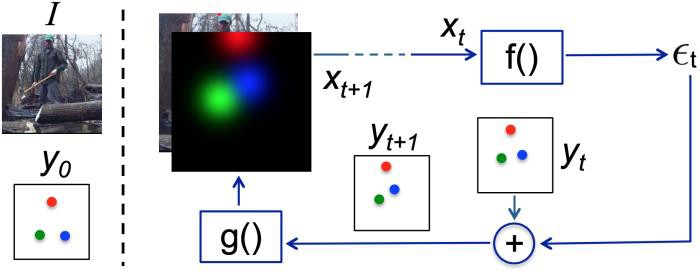}
	\caption{An implementation of Iterative Error Feedback (IEF). \cite{carreira2016human}} 
	\label{fig11}
\end{figure}
Another framework for estimating poses is proposed later in 2016, called Stacked Hourglass Network \cite{newell2016stacked}. The architecture is designed to capture and consolidate image features across all scales by processing steps of pooling and up-sampling to get the final set of predictions. To address localization error problem and the redundant detection problem, which cannot be handled by previous approaches, \cite{fang2017rmpe} published a framework called RMPE (Regional Multi-person Pose Estimation) in Dec, 2016. RMPE is composed of three components. The first component is “SSTN (Symmetric Spatial Transformer Network) and parallel SPPE (Single Person Pose Estimator)”, and SSTN includes STN (Spatial Transformer Network) and SDTN (Spatial De-transformer Network) and SPPE. The former is the forward procedure and inverse procedure to extract high-resolution dominant human proposals to SPPE, then remap to original human proposal image, and remap the estimated pose to the original image coordinate. The latter is used to enhance SPPE when given imperfect human proposals. The second component is parametric pose NMS (Non-Maximum Suppression) used for redundant proposals elimination. The last component is PGPG (Pose-guided Proposals Generator) for enriching existing training samples. \cite{papandreou2017towards} proposed a method called G-RMI for multi-person detection and 2-D pose estimation, which had an excellent outperformance on the challenging COCO keypoints task in 2017. The model estimates the pose by locating the body keypoints based on activation heatmaps and offsets predicted by fully convolutional ResNet. The combination of the output from detection and estimation part is processed by a novel aggregation procedure to obtain highly localized keypoint predictions. The author also replaced the cruder box-level NMS by a novel form of keypoint-based NMS, and box-level scoring by a novel form of keypoint-based confidence score estimation to avoid duplicate pose detections. In the same year later, an excellent method called Mask R-CNN \cite{he2017mask}, which is proposed by Facebook AI Research (FAIR), estimates human poses by generated one-hot mask. The location of each keypoint is modelled as this maks, and Mask R-CNN predicts K masks, one for each of K keypoint types (e.g, neck, head). Although Mask R-CNN made a great progress, it still cannot handle several challenges such as occluded keypoints, invisible keypoints, and crowded back-ground. The main reasons are these “hard” joints being hard to predict based on the appearance features like the torso point, and the imperfection during the training to address these “hard” joints. To solve above problems, in April, 2018, CPN (Cascaded Pyramid Network) \cite{chen2018cascaded} for multi-person pose estimation is published by Tsinghua University and HuaZhong University of Science and Technology. The CPN network architecture is two-stage framework, which is GlobalNet for relative keypoints estimation, and RefineNet for mining loss by online hard keypoint.
\begin{figure}[!htp]
	\centering
	\includegraphics[scale=0.45]{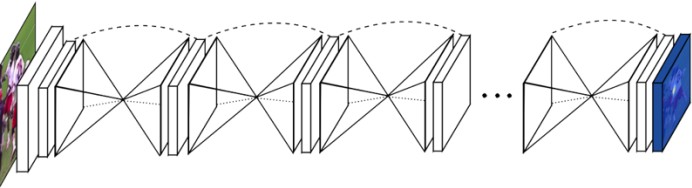}
	\caption{Structure overview of stacked hourglass networks.\cite{choi20213dcrowdnet}} 
	\label{fig12}
\end{figure}
\begin{table*}[!htp]
	\footnotesize
	\centering
	\caption{Overview of 2D datasets}
	\label{tab1}
	\begin{tabular}{|l|l|l|l|}
		\hline
		\textbf{Title} &  \textbf{Method} & \textbf{Network Combination} & \textbf{Year} \\ \hline
		Pose estimation and behavior classification of broiler chickens based on deep neural networks & RMPE & Faster R-CNN + Hourglass & 2021 \\ \hline
		Human pose estimation with iterative error feedback & IEF & Detector + IEF & 2016 \\ \hline
		Stacked hourglass networks for human pose estimation &  Hourglass Network & Stacked Hourglass Network  & 2016 \\ \hline
		Towards accurate multi-person pose estimation in the wild &  G-RMI & Faster R-CNN+ G-RMI  & 2017  \\ \hline
		Mask R-CNN &  Mask R-CNN & Faster R-CNN + ResNet-FPN  & 2017  \\ \hline
		Cross-view tracking for multi-human 3d pose estimation at over 100 fps &  CPN & FPN + CPN  & 2020  \\ \hline	
	\end{tabular}
\end{table*}

Generally, top-down approaches are simply the combination of existing object detectors and single human pose estimator. Table \ref{tab1} describes these combinations for each mentioned human pose estimation method.
Although most of top-down methods achieved high accuracy performance, they face the real-time speed problem caused by the complex computation in the network architecture.
\subsection{Iterative Error Feedback}
IEF framework \cite{carreira2016human} is designed to extend the power of Convolutional Networks to include both input and output spaces by introducing a top-down feedback mechanism. This mechanism allows model to predict the error and correct it at current estimation time, instead of predicting the target output directly.
Figure \ref{fig11} modifies the implementation of IEF for 2D human pose estimation. On the left side, the input includes the image denoted as $I$, and a set of 2D points presenting the initial guess of keypoints denoted as $Y_0$. Only 3 out of 17 key pointsare visualized in the figure which are the right wrist (green), left wrist (blue), and top the of head (red). The input is defined as $X_t = I \oplus g(y_{t-1}) $, where $I$ is representative for the image and $y_{t-1}$ is the previous output. Function $f$, modeled here as a ConvNet, receives the input $x_t$, then outputs a correction $\epsilon_t$. The current correction is passed into output $y_t$, leading to the new keypoint position estimates $y_{t+1}$. Function $g$ renders the new keypoint, which is stacked with image $I$, to output the $x_{t+1}$. The main role of function $g$ is to convert 2D keypoint position into Gaussian heatmap channel one by one so that it can be part of the input with the image for the next iteration. The entire workflow is repeated $T$ time until obtaining a refined $y_{t+1}$ meaning the prediction gets closer to the ground truth. The mathematical background of the procedure can be described by the following equations:
\begin{equation}
	\epsilon_t = f(x_t)
\end{equation}
\begin{equation}
	y_{t+1} = y_t + \epsilon_t
\end{equation}
\begin{equation}
	x_{t+1} = I \oplus + g(y_{t+1})
\end{equation}

where function $f$ has additional learned parameter as $\theta_f$, and $g$ has additional learned parameter $\theta_g$. These parameters are learned by optimizing the equation below,
\begin{equation}
 \min_{\nabla_{t=0}^T} h(\epsilon_t, e(y,y_t))	
\end{equation}
where $\epsilon_t$ is the predicted output and $e(y,y_t)$ is target corrections. The function $h$ is distance measurement, such as $q$ quadratic loss. $T$ represents for correction steps number taken by the model. The author experimented between IEF framework and several benchmark datasets for 2D human pose estimation: the MPII Human Pose dataset, and Leeds Sports Pose dataset (LSP). The using evaluation metric is PCKh$@0.5$.
\begin{figure}[!htp]
	\centering
	\includegraphics[scale=0.45]{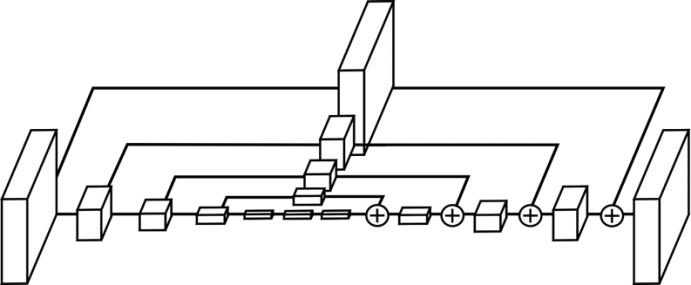}
	\caption{A single "Hourglass" module illustration. \cite{choi20213dcrowdnet}} 
	\label{fig13}
\end{figure}
\subsection{Stacked Hourglass Network}
\begin{figure}[!htp]
	\centering
	\includegraphics[scale=0.45]{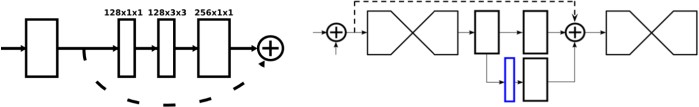}
	\caption{Left: residual module. Right: intermediate supervision process illustration. \cite{choi20213dcrowdnet}} 
	\label{fig14}
\end{figure}

Stacked Hourglass Networks \cite{choi20213dcrowdnet} is designed to capture features across scale over image, which is composed of two steps
– pooling and up-sampling. Therefore, the framework of this method, as shown in figure \ref{fig12}, is the symmetric distribution of capacity between bottom-up (from high resolution to low resolution using pooling) and top-down (from low resolution to high resolution using up-sampling) to improve the network performance.
The figure \ref{fig13} illustrates the a single “hourglass” module, which presented that each box in the figure referring to a residual module (figure \ref{fig14}). In whole hourglass, the number of features stays consistent. While maintaining the overall hourglass shape, this residual learning, which is constructed with three ConvNet having distinct scales in which batch normalization and ReLu inserted between them, extracts high-quality image features. The second path skips the path and has only one kernel, which is a convolutional layer having a scale of 1. During the process, there is no change with the data in addition to the data depth. Generally, the concept of this single “hourglass” module in figure 13 is an encoder and decoder architecture. The module down-sample the image feature first, followed by up-sampling the features to generate a heat-map.

Author applied a fourth-order residual module for each hourglass module to extracts features from the original scale to the 1/16 scale. The setup of the hourglass network starts by down- sampling features to a very low resolution. The network branches off and applies more convolutions at the original pre-pooled resolution, which takes place at each max pooling step. After obtaining the lowest resolution features, the network begins upsampling and gradually combining feature information of various scales. The nearest neighbor up-sampling method is used for lower resolution.
Stacked Hourglass Networks was tested with 2 benchmark datasets, which are FLIC and MPI. Regarding the FLIC dataset, the achievement was 99\% PCK@0.2 accuracy on the elbow and 97\% on the wrist. The test with the second dataset resulted in an average of 3.5\% (PCKh@0.5) and 12.8\% error for the prediction task with difficult joints like the wrist, elbows, knees, and ankles. The final result for elbow is 91.2\% accuracy, and wrist is 87.1\% accuracy.
\subsection{Regional Multi-Person Pose Estimation (RMPE)}
\begin{figure}[!htp]
	\centering
	\includegraphics[scale=0.45]{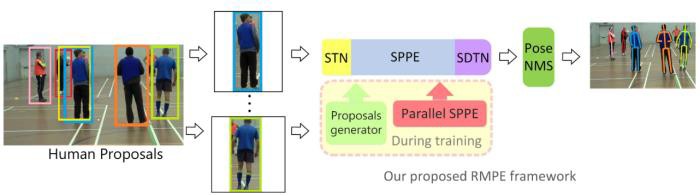}
	\caption{Pipeline of RMPE framework. \cite{fang2017rmpe}} 
	\label{fig15}
\end{figure}

RMPE is composed of three components \cite{fang2017rmpe} which are SSTN, Parametric Pose NMS, and PGPG. The pipeline is shown in the figure \ref{fig15}, which starts by feeding the human bounding boxes generated from detector into the “Symmetric STN + SPPE” (SSTN) model to generate pose proposals. After that the author Parametric Pose NMS to refine produced pose proposals in order to estimate human poses. Additionally, a module called “Parallel SPPE” is attached with the main module SSTN to avoid local minimums and extend the power of SSTN. For existing training examples enrichment, PGPG is proposed.
\subsubsection{SSTN and Parallel SPPE}
\begin{figure*}[!htp]
	\centering
	\includegraphics[scale=0.9]{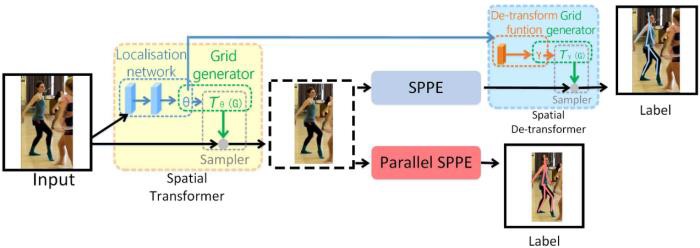}
	\caption{The illustration of SSTN + parallel SPPE module. \cite{fang2017rmpe}} 
	\label{fig16}
\end{figure*}

In fact, SPPE performs well only with single person image and very sensitive to localization errors. The symmetric STN + parallel SPPE is used to improve the SPPE's ability to deal with imperfect human proposals. The module process is visualized in the figure \ref{fig16}.

STN and SDTN are spatial transformer networks and spatial de-transformer networks that achieve excellent performance in automatically selecting RoIs. The STN is a forward procedure, which extracts high-resolution dominant human proposals, followed by passing into SPPE. As a result, the pose from that process is mapped into the original human proposal image. In mathematics, the STN performance can be modified as
\begin{equation}
\begin{pmatrix}                     
x_i^s\\
y_i^s
\end{pmatrix}            =    \begin{bmatrix}
			     \theta_1 & \theta_2 & \theta_3
				\end{bmatrix}          \begin{pmatrix}
							x_i^s\\
							y_i^s\\
							1
							\end{pmatrix}
\end{equation}
where $\theta_1$, $\theta_2$, $\theta_3$ are vectors in $\mathbb{R}^2$, and $x_i^s$, $y_i^s$ are coordinates before the transformation, and $x_i^t$, $y_i^t$ are coordinates after the transformation.
The SDTN is an inverse procedure used to remap the estimated output to the original image coordinates. In term of mathematical description, the SDTN computes $\gamma$ for the inverse transformation and produce grids based on $\gamma$:
\begin{equation}
	\begin{pmatrix}                     
		x_i^s\\
		y_i^s
	\end{pmatrix}            =    \begin{bmatrix}
		\gamma_1 & \gamma_2 & \gamma_3
	\end{bmatrix}          \begin{pmatrix}
		x_i^s\\
		y_i^s\\
		1
	\end{pmatrix}
\end{equation}
The following equation is obtain because SDTN is an inverse procedure of STN:
\begin{equation}
\label{eq7}	
\begin{bmatrix}
	\gamma_1 & \gamma_2 
\end{bmatrix}   =                 {\begin{bmatrix}
\theta_1 & \theta_2
\end{bmatrix}}^{-1}
\end{equation}

\begin{equation}
	\label{eq8}
	\gamma_3 = -1 \times \begin{bmatrix}
		\gamma_1 & \gamma_2 
	\end{bmatrix}   =    \theta_3           
\end{equation}
For back-propagation through SDTN, $\frac{\partial(W,b)}{\partial\theta}$ can be derived as:
\begin{equation}
\frac{\partial(W,b)}{\partial[\theta_1 \theta_2]} = \frac{\partial(W,b)}{\partial[\gamma_1 \gamma_2]} \times \frac{\partial[\gamma_1 \gamma_2]}{\partial[\theta_1 \theta_2]}  + \frac{\partial(W,b)}{\partial\gamma_3} \times \frac{\partial \gamma_3}{\partial [\gamma_1 \gamma_2]} \times \frac{\partial[\gamma_1 \gamma_2]}{\partial[\theta_1 \theta_2]}
\end{equation}
with respect to $\theta_1$ and $\theta_2$, and
\begin{equation}
\frac{\partial(W,b)}{\partial\theta_3} = \frac{\partial(W,b)}{\partial\gamma_3} \times \frac{\partial \gamma_3}{\partial \theta_3}
\end{equation}
with respect to $\theta_3$, and $\frac{\partial\gamma_3}{\partial\theta_3}$ can be derived from eq \ref{eq7} and \ref{eq8}. 
Parallel SPPE is attached to improve the human dominant regions extractor during the training time. This branch shares the
same STN in comparison with original SPPE in addition to SDTN, which is discarded.

\subsubsection{Parametric Pose NMS}
To solve the problem of redundant pose estimations, the author applied Parametric Pose NMS being similar to the previous subsection, the pose $P_i$ with $m$ joint is defined as $\{(k_i^1,c_i^1),.....,(k_i^m,c_i^m)\}$, where $k_i^j$, $c_i^j$ are the $j^{th}$location and confidence score of joints respectively.
In term of elimination criterion, the author defined a pose distance metric $x(P_i, P_j|\Lambda)$ to measure and evaluate the pose similarity, and a threshold $\eta$ for elimination task. $\Lambda$ is a parameter set of function $d(.)$. The function can be written as below:
\begin{equation}
f(P_i,P_j|\Lambda,\eta) = \mathbb{I}[(P_i,P_j|\Lambda, \lambda)\leq \eta]
\end{equation}
if $d(.)$ is less than $eta$, the output of $f(.)$ should be 1, which describe that the pose $P_i$ should be eliminated due to redundancy with reference pose $P_j$.
\subsubsection{PGPG}
The aim of PGPG is to augment data so that the data is enriched for the two-stage pose estimation. As a result, the model SSTN + SPPE adapts to the imperfect human proposals. RMPE obtained 76.7 mAP on the MPII dataset, and managed to handle inaccurate bounding boxes and redundant detections.
\subsection{G-RMI}
\begin{figure}[!htp]
	\centering
	\includegraphics[scale=0.45]{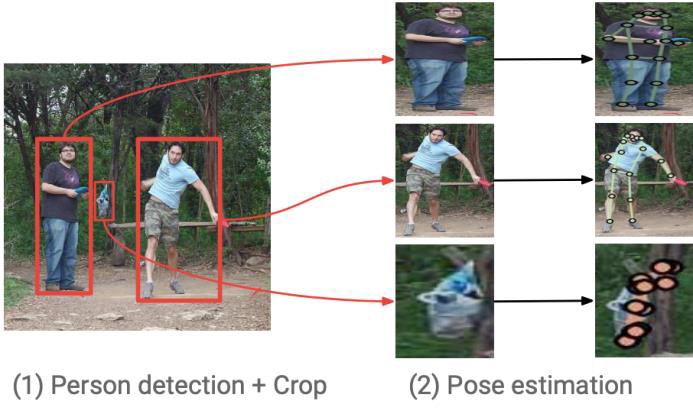}
	\caption{G-RMI Framework overview. \cite{papandreou2017towards}} 
	\label{fig17}
\end{figure}

G-RMI \cite{papandreou2017towards} estimates human poses by predicting dense heatmaps and offset using a fully convolutional ResNet. The overview of G-RMI model is presented in the figure \ref{fig17}, which is separated into 2 parts. The first part employed a Faster R-CNN human detector to generate bounding boxes. The second section uses an estimator to localize keypoints and rescore the relevant human proposal in the image crop extracted around each person.

Essentially, this model is a classification and regression approach that first classifies whether it is in the vicinity of each of the K keypoints or not (called a "heatmap"), then predicts a 2D local offset vector to achieve a more precise estimate of body keypoint location.

\subsubsection{Image Cropping}
At the beginning, without distorting the image aspect ratio, the author adjusted either the width or the heigh of the returned bounding box from the detector to make all boxes have the same fixed aspect ratio. Afterwards, during training, a rescaling factor equaling to 1.25 and a random rescaling factor between 1.0 and 1.5 are applied to enlarge the boxes to include the background context. Finally, before setting the aspect ratio value to 353/257 = 1.37, the resulting box is cropped and resized to the fixed parameter with 353 pixels for height an 257 pixels for width.
\subsubsection{Heatmap and Offset prediction}
On a cropped image, a ResNet-101 backbone network is used to generate heatmaps (one channel per keypoint) and offset (two channels presenting x and y-directions respectively per keypoint) for a total output of $3k$ channels, where $K=17$ is the number of keypoints. The using model architecture is ImageNet pre-trained ResNet-101 with the replacement of the last layer by $1\times1$ convolution. Atrous convolution is used to have a stride of 8. Bilinearly up-sampled is applied to enlarge the network output back to the $353 \times 257$ crop size. To compute the heatmap probability, the author proposed that $h_k(x_i) = 1$ if $||x_i-k_k|| \leq R$ to compare with the radius R, where $h_k(x_i)$ is the probability that the point $x_i$ is within a disl of radius R from the location $l_k$ of the $k-th$ keypoint. $h_k(x_i)=0$ if the keypoint is outside R. After creating the heatmaps and offsets, the author presented a method for combining them to create highly localized activation maps:
\begin{equation}
	f_k(x_i) = \sum_{j}^{max}\frac{1}{\pi \mathbb{R}} G(x_j + F_k(x_i) - x_i)h_k(x_i)	
\end{equation}
\subsubsection{OKS-Based NMS}
The author applied standard approach to measure overlap using IoU of the boxes, and redundant boxes are removed. The overlap measurement in G-RMI is processed with object keypoint similarity (OSK). To filter highly overlapping boxes, a relatively high IOU-NMS threshold (0.6) is applied at the output of the box detector. G-RMI won the 2016 COCO keypoints challenge with an average precision of 0.649 on the COCO test-dev set and 0.643 on the test-standard set.

\begin{figure}[!htp]
	\centering
	\includegraphics[scale=0.65]{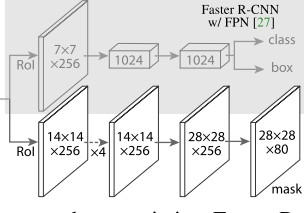}
	\caption{The implementation of ResNet-FPN for human pose estimation. \cite{papandreou2017towards}} 
	\label{fig18}
\end{figure}
\subsection{Mask R-CNN}
Mask R-CNN \cite{he2017mask} includes both human detector and pose estimator in its framework. For the human pose estimation, the keypoint’s location is structured as a one-hot mask, and adopt Mask R-CNN to estimate K masks with one for every K keypoint type. A minor modification is made when adapting segmentation to keypoints. The expected training output for each keypoint of an instance on an image is a one-hot \(m \times m\) binary mask with only a single pixel labeled as foreground. The author minimizes the cross-entropy loss over an m2-way softmax output for single visible ground-truth keypoint. To be noted that K keypoints are still treated independently, even though the mode is instance segmentation. The keypoint head architecture is shown in the figure 18, which is built by adopting the ResNet-FPN variant. Mask-RCNN scored 62.7 \(AP^{kp}\), which is 0.9 point higher than the winner of the COCO 2016 keypoint detector competition, which employs a multi-stage processing pipeline.

\subsection{Cascaded Pyramid Network (CPN)}
CPN \cite{chen2018cascaded} for multi-person pose estimation model is built to deal with current challenging problems called “hard keypoint”. This problem is about occluded keypoints, invisible keypoints and the complicated background etc. The CPN network architecture is two-stage framework, which is GlobalNet for relative keypoints estimation, and RefineNet for mining loss by online hard keypoint.
\begin{figure*}[!htp]
	\centering
	\includegraphics[scale=0.95]{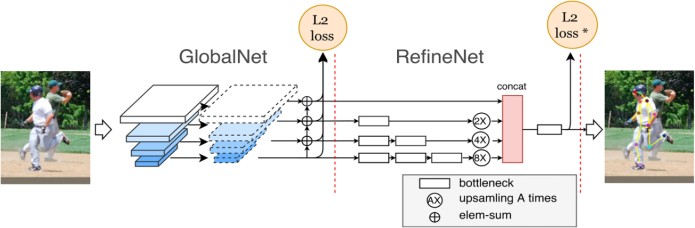}
	\caption{The network structure of Cascaded Pyramid Network (CPN). \cite{chen2018cascaded}} 
	\label{fig19}
\end{figure*}
In CPN, the visible keypoints are easy to identify because the fixed shape, which helps in obtaining texture information to easily get contextual information around the joint’s location. For example, nose, left elbow, and right hand are visible easy keypoints. Another kind keypoint is hard visible one, which is obscured by clothes such as left knee, right knee, and left hip. In addition, other joints are hidden and tough to predict such as the right shoulder in the figure \ref{fig19}. Increasing the local receptive field can help in refining the information context. As a result, CPN easily classified the human body joints into simple parts and difficult parts.
\subsubsection{GlobalNet}
GlobalNet \cite{chen2018cascaded} basically is a forward CNN, which is implemented to detect easy visible human keypoints like eyes, arm, and other easy to detect parts.
\subsubsection{RefineNet}
RefineNet \cite{lin2017refinenet} is designed to detect invisible hard keypoints by integrating multiple receptive fields information with the feature maps of the pyramid model generated by GlobalNet. Finally, all feature maps, having the same size, are concatenated so that a correction for ambiguous keypoints is achieved. To mine the difficult keypoints, first RefineNet concatenates when using features of multi-layers, and the second thing is online hard keypoints missing techonology for the second-level network. In summary, RefineNet's work is based on combining low-level and high-level features via convolution operations. CPN achieved average precision of 73.0 on the COCO test-dev dataset and 72.1 on the COCO test-challenge dataset using the COCO keypoint benchmark, a 19\% improvement over the COCO 2016 keypoint challenge of 60.5.

\section{Evaluation metrics}
\label{sec:EXP}
Each dataset has its own features and evaluation metrics. The following summary presents different evaluation metrics.
\subsection{Percentage of Correct Parts (PCP)}
This metric is commonly used in preliminary research, which describes the localization accuracy of limbs. If the limb's two endpoints are within a certain distance of the corresponding ground truth endpoints, it is precisely localized. The threshold can be set to 50\% of the length of the limb.
\subsection{Percentage of Correct Keypoints (PCK)}
This metric measure and evaluate the body joints-level localization accuracy. A predicted body joint is considered correct if it falls within the ground-truth joint's threshold pixels. The threshold can be set as a percentage of the person's bounding box size, a pixel radius normalized by the torso height of each test example, or 50\% of the length of the head segment of each test image.
\subsection{The Average Precision (AP)}
In AP measurement scheme, the predict joint is considered as true positive, if the joint is within a certain distance of the ground-truth joint For each keypoint, the connection between anticipated joint and ground-truth poses is determined individually. In the case of multi-person pose evaluation, all prediction result are assigned to reference poses one by one according to the PCKh score order, whereas false positive outputs are not assigned. Average Precision (AP), Average Recall (AR), and variants of these terms: The AP, AR metric is used to evaluate multi-person pose estimation and is based on a similarity measure: The same capability as Intersection over Union is Object Keypoint Similarity (OSK) (IoU). AP/AR is also reported in the COCO dataset, with different human body scales.

\begin{figure*}[!htp]
	\centering
	\includegraphics[scale=0.45]{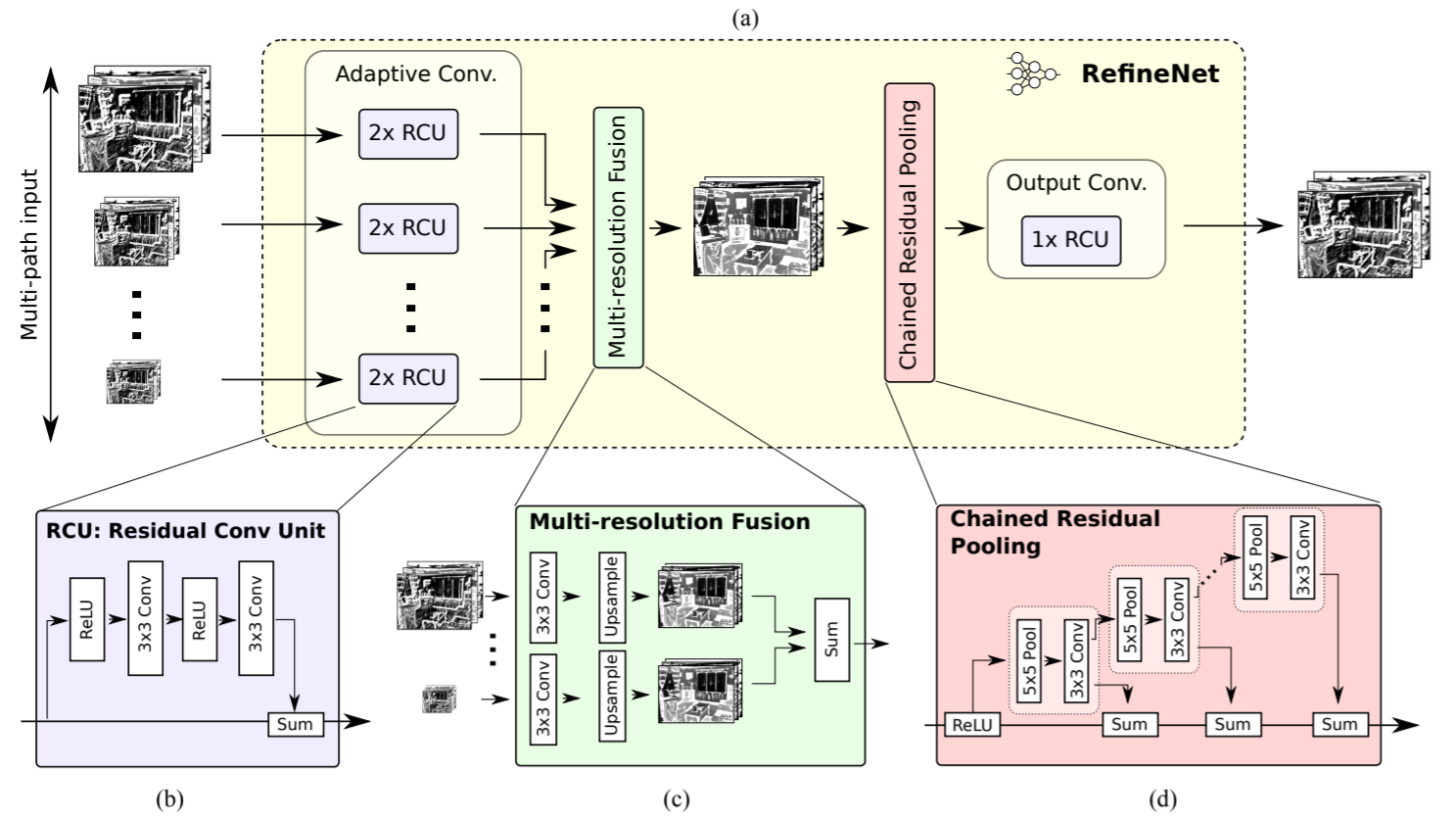}
	\caption{RefineNet components use residual connections with identity mappings. Gradients can thus be directly propagated within RefineNet via local residual connections, as well as to the input paths via long-range residual connections, resulting in effective end-to-end training of the entire system. Courtesy \cite{lin2017refinenet}
} 
	\label{fig20}
\end{figure*}
\section{Conclusion}
\label{sec:CON}
The purpose of this paper is to provide researchers with an extensive review of deep learning methods-based 2D images for human pose estimation, which have only focused on top-down approaches since 2016. Approaches are categorized into two types of methods: the two-step framework (top-down approach) and the part-based framework (bottom-up approach). The two-step framework first incorporates a person detector in the bounding boxes, then predicting the pose within each box independently. In the part-based framework, the system starts detecting all body parts in the image, and then associate parts belonging to distinct persons. This paper presents significant detectors and estimators depending on mathematical background, the challenges and limitations, benchmark datasets, evaluation metrics, and comparison between methods.



{\small
\bibliographystyle{IEEEbib}
\bibliography{tt}
}
\end{document}